\documentclass{article}

\usepackage[final]{corl_2019} % Uncomment for the camera-ready ``final'' version
\usepackage{amssymb}
\usepackage{booktabs}
\usepackage{graphicx}
\usepackage{subfigure}
\usepackage{caption}
\usepackage{bbm}
\usepackage{color}
\usepackage{amsmath}

\usepackage{amssymb}
\usepackage[normalem]{ulem}
\usepackage[labelfont=bf,font=small]{caption}
\usepackage{algorithm}
\usepackage{algorithmicx}
\usepackage{algpseudocode}
\usepackage{graphicx}
\usepackage{wrapfig}
\begin{document}

\title{Nonverbal Robot Feedback for Human Teachers}

% The \author macro works with any number of authors. There are two
% commands used to separate the names and addresses of multiple
% authors: \And and \AND.
%
% Using \And between authors leaves it to LaTeX to determine where to
% break the lines. Using \AND forces a line break at that point. So,
% if LaTeX puts 3 of 4 authors names on the first line, and the last
% on the second line, try using \AND instead of \And before the third
% author name.

% NOTE: authors will be visible only in the camera-ready (ie, when using the option 'final'). 
% 	For the initial submission the authors will be anonymized.

\author{
Sandy H. Huang\thanks{Equal contribution.}\, \thanks{Now at DeepMind.} \,, Isabella Huang$^*$, Ravi Pandya$^*$, Anca D. Dragan\\
  University of California, Berkeley\\
}

\newcommand{\adnote}[1]{
 {\textcolor{blue}{\textbf{[#1 --AD]}}}}

\newcommand{\SH}[1]{
 {\textcolor{cyan}{\textbf{[#1 --SH]}}}}
 
\newcommand{\IH}[1]{
 {\textcolor{magenta}{\textbf{[#1 --IH]}}}}
 
 \newcommand{\RP}[1]{
 {\textcolor{red}{\textbf{[#1 --RP]}}}}

% Labels in IEEE format
% Equation
\newcommand{\eref}[1]{(\ref{#1})}
% Section
\newcommand{\sref}[1]{Sec. \ref{#1}}
% Figure
\newcommand{\figref}[1]{Fig. \ref{#1}}
%Algorithm
\newcommand{\algoref}[1]{Algo. \ref{#1}}
\newcommand{\prg}[1]{\noindent\textbf{#1. }} 
\newcommand{\be}[1]{\textbf{\emph{#1}}}
\newcommand\Tstrut{\rule{0pt}{2.6ex}}         % = `top' strut
\newcommand\Bstrut{\rule[-0.9ex]{0pt}{0pt}}   % = `bottom' strut

\maketitle

%===============================================================================

\begin{abstract}
Robots can learn preferences from human demonstrations, but their success depends on how informative these demonstrations are. Being informative is unfortunately very challenging, because during teaching, people typically get no transparency into what the robot already knows or has learned so far. In contrast, human students naturally provide a wealth of nonverbal feedback that reveals their level of understanding and engagement. In this work, we study how a robot can similarly provide feedback that is minimally disruptive, yet gives human teachers a better mental model of the robot learner, and thus enables them to teach more effectively.
Our idea is that at any point, the robot can indicate what it thinks the correct next action is, shedding light on its current estimate of the human's preferences. We analyze how useful this feedback is, both in theory and with two user studies---one with a virtual character that tests the feedback itself, and one with a PR2 robot that uses gaze as the feedback mechanism. We find that feedback can be useful for improving both the quality of teaching and teachers' understanding of the robot's capability.
\end{abstract}


%===============================================================================

\section{Introduction}
If robots are to be useful to humans, they need to do more than optimize reward functions---they need to be able to figure out what reward functions should be optimized in the first place.
Inverse Reinforcement Learning (IRL)~\cite{Ng_2000} enables robots to infer preferences from human demonstrations. For instance, by collecting data of human drivers, we can infer a human-like driving style~\cite{Abbeel_2004,Levine_2012}. Traditionally, IRL is applied in settings where data is collected offline from people who have no idea that a robot is supposed to learn from this data. When talking about learning preferences for the purpose of assisting people---which is arguably the goal for most robots and AI agents---there is an opportunity to explicitly involve people in teaching the robot about what they want. It can be much more effective for me to actively teach my robot how to organize my kitchen, for instance, instead of having the robot collect data of me putting things away over and over again until it eventually figures out my preferences. When I teach, I get the chance to select examples that might be especially informative to the robot---ones that effectively illustrate the core of my approach.

Unfortunately, effective teaching is tricky even when we teach other people. We have to figure out what the person knows and does not know, what teaching strategy works best for them as an individual, etc. Teaching robots is monumentally harder. We have much poorer mental models of how robots learn compared to our mental models of how humans do. What is more, when we teach humans we receive a great deal of feedback from them.
One traditional way we get feedback is through tests, by asking questions to probe the learner's understanding. More interestingly though, human students continually provide (nonverbal) feedback to the teacher \emph{during the teaching process itself}. They look confused or bored, nod along, fidget, or gaze at various things~\cite{Angelo_1993,Webb_1997}.
\begin{wrapfigure}{R}{0.43\textwidth}
    \centering
    \includegraphics[width=0.42\textwidth]{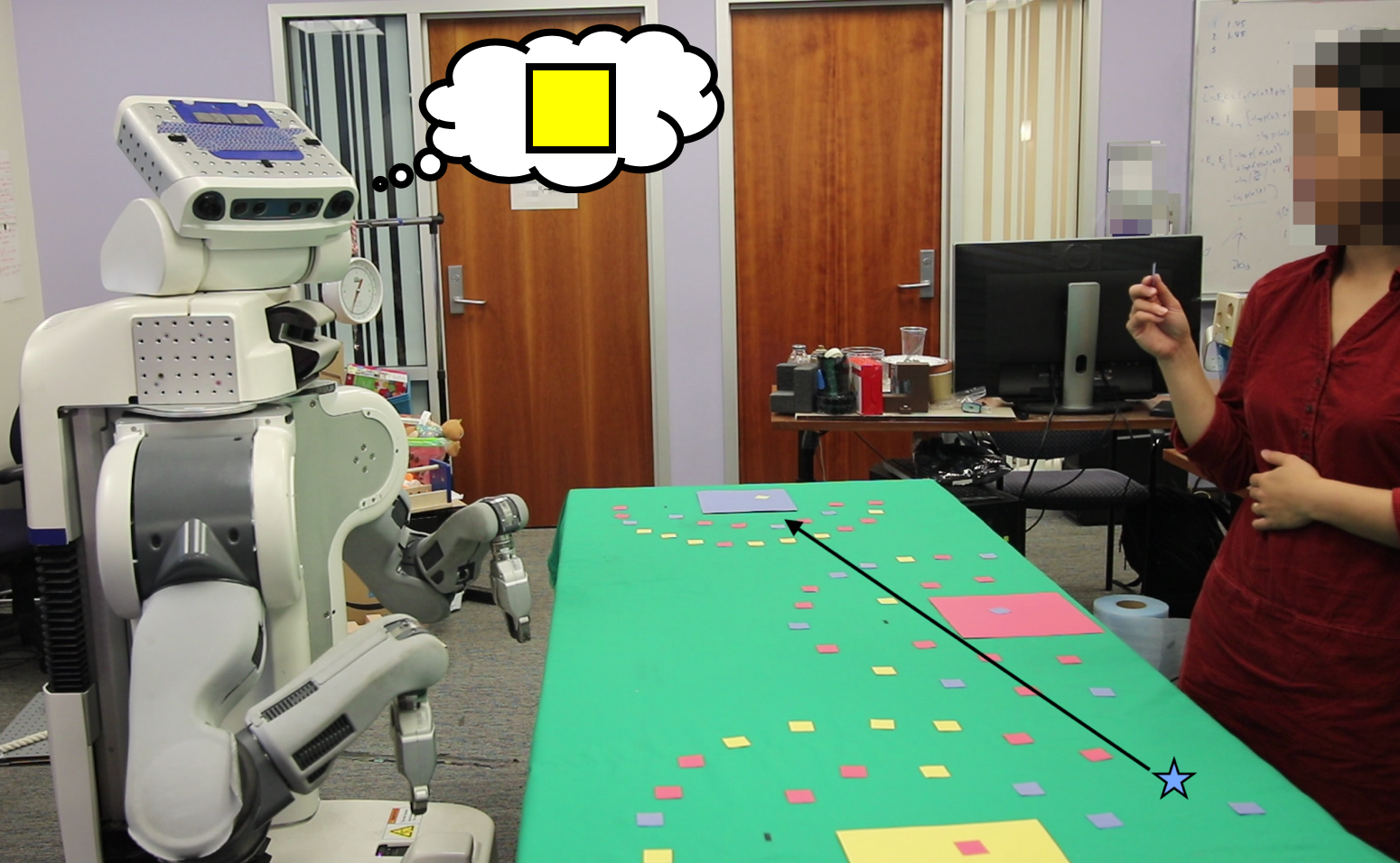}
    \caption{The robot uses its learned preferences to anticipate the teacher's next action (i.e., placing the object in the yellow bin), and then uses  gaze to communicate its belief to the teacher.}
    \label{fig:frontfig}
\end{wrapfigure}

Our goal is for robots to provide similar feedback, at the same time as the teacher is providing examples. We take a step towards that in this paper, based on a simple idea: to use the states that the teacher is in as opportunities to inform the teacher about what the robot expects them to do, according to the robot's current understanding of the task.
We resort to \emph{nonverbal cues} (e.g., gaze) as an intuitive and minimally-disruptive way for the robot to signal how it expects the teacher to act next.
For example, imagine teaching the robot how to organize objects in a decluttering task (\figref{fig:frontfig}). Every time you pick up an object, the robot gazes at where it thinks the object should go. This helps you gain a better understanding of the robot's current hypothesis about your preferences, which helps you figure out a better next example to provide. It also helps you realize when the robot has finished learning, and you can stop teaching.

Our main contributions are as follows: 1) we propose a form of feedback that robots can provide while teaching is ongoing, that consists of a prediction of the teacher's next action along with the confidence; 2) we provide a theoretical motivation for why this form of feedback should improve teaching effectiveness, by introducing an algorithmic teaching model that takes feedback into account; and 3) we test out this feedback with real people, both in an online user study with virtual learners and in an in-person study with a real robot, where gaze is the feedback mechanism. We find that feedback can improve the teacher's estimate of the learner's understanding, quantifiably change the teacher's strategy, and result in higher learner accuracy. Results with real users support our theoretical analysis. With gaze, we find that these effects are stronger when participants know explicitly about the feedback they should expect---otherwise, some participants interpret gaze differently, e.g., as acknowledgement or as purely functional.
\section{Related Work}
\noindent\textbf{Improving quality of human input.}
Teaching robots via demonstration~\cite{Argall_2009} is challenging; humans may have trouble providing useful demonstrations~\cite{Cakmak_2010, Khan_2011} and knowing when to stop teaching~\cite{Sena_2018}.
Robots can take a more active role in learning, by asking for additional demonstrations where they are uncertain~\cite{Inamura_1999,Shon_2007,Chernova_2009,Silver_2012} or know information is missing~\cite{Miura_2005}, or by asking clarification questions~\cite{Whitney_2017}. Robots can ask for different kinds of teacher input as well---e.g., labels, feature queries, or demonstrations---to maximize usefulness~\cite{Cakmak_2012}. However, when robots are active learners, humans lose control over the teaching process, which can make them feel frustrated and disengaged~\cite{Chao_2010,Cakmak_2010b}.

We take an orthogonal approach, in which humans maintain control of teaching, while the robot tries to be \emph{transparent} about what it has learned. Transparency enables \emph{algorithmic teaching}, in which the teacher's understanding of how the learner learns enables her to select teaching examples that optimally teach this learner.\footnote{Since the teacher knows the task, algorithmic teaching is typically more efficient than active learning~\cite{Zhu_2015}.} Robot learners can be more transparent by demonstrating their current learned policy~\cite{Calinon_2007}, allowing teachers to ask questions that probe their understanding~\cite{Steels_2001,Lim_2014},
or showing where they succeed and fail~\cite{Sena_2018}. However, these approaches require people to stop teaching and separately test the robot. In contrast, we focus on how robots can provide feedback \emph{at the same time} as humans are teaching---similar to the nonverbal feedback cues that human students give. This form of feedback requires minimal context switching from the teacher and does not add to the total interaction time. On the real robot, we implement this feedback as gaze.

\noindent\textbf{Robot gaze.}
Robots can use gaze to be transparent in social interactions with humans (see~\citet{Admoni_2017} for a detailed survey). Robot gaze can also be useful in human-robot collaboration tasks, by disambiguating referenced objects~\cite{Admoni_2016,Boucher_2012}, communicating which action the robot is about to take~\cite{Moon_2014}, and influencing the human's actions~\cite{Admoni_2014}.

In this work, we also use robot gaze to communicate to humans, but specifically while the robot is \emph{learning}. \citet{Thomaz_2008} explored this for reinforcement learning (RL) agents: when agents use gaze to communicate what action they are about to take, humans are able to provide more informative rewards, thus speeding up learning. However, in RL the robot acts while the person observes.
When robots instead learn from human demonstrations, the opposite is true, which makes the robot's learning opaque. Taking inspiration from the RL setting, we aim at having the robot always convey what it thinks is the optimal action to take, so that the person can adjust their demonstrations appropriately.
\section{Nonverbal Feedback}

\subsection{Assumptions on robot learning algorithm}
We focus on robot learners that infer \emph{reward functions} from demonstrations via IRL~\cite{Ng_2000}. The benefit of IRL is that this underlying reward function typically generalizes better across tasks, compared to directly learning a mapping from states to actions, as in behavior cloning~\cite{Pomerleau_1991,Ross_2011}. In addition, humans are naturally inclined to infer the objectives of other agents~\cite{Baker_2009,Jara_2016}, and thus might expect robot learners to do the same. We parameterize the reward function $\mathcal{R}_\theta$ as a linear combination of reward features $\phi(\cdot)$ with weights $\theta$,
   $ \mathcal{R}_\theta(s, a) = \theta^\top \phi(s, a) $, 
where $s$ is the state and $a$ is the action. There is no limitation on what these reward features can be, so this assumption is not restrictive~\cite{Abbeel_2004}.

The robot maintains a belief $b(\theta)$ over reward function parameters. We model humans as providing demonstrations $(s,a)$ that are approximately optimal, according to a Boltzmann distribution~\cite{Ortega_2016}. This induces the following observation model that links human actions to the reward parameters,
\begin{equation}
    p(a|s,\theta)\propto e^{\beta Q^*_\theta(s, a)}\, ,
    \label{eqn:softmax}
\end{equation}
where $\beta$ specifies the level of suboptimality and $Q^*_\theta(s,a)$ is the action-value function: the discounted sum of future rewards, after taking action $a$ in state $s$ and acting optimally thereafter with respect to $\mathcal{R}_\theta$.
As in Bayesian IRL~\cite{Ramachandran_2007}, the robot uses this observation model to update its belief over $\theta$:
\begin{equation}
    b'(\theta) \propto p(a | s, \theta) \, b(\theta) \, .
    \label{eqn:bayes_irl}
\end{equation}

\subsection{Generating feedback}
We propose and investigate a form of feedback where at every step, the robot communicates what it believes the optimal action is. Intuitively, this should help human teachers understand what the robot knows and does not know as they proceed through teaching the task, enabling them to adapt what they teach and recognize when they can stop teaching. In addition, this form of feedback is minimally disruptive and allows teachers to maintain control of teaching, in contrast to tests of comprehension and active learning, respectively. We leave open the question of how this feedback should be communicated; we experiment with movement of a virtual avatar and gaze on a real robot.

\noindent\textbf{Feedback target.}
The robot first uses its current belief over reward parameters to predict the human's next action, using the observation model from \eref{eqn:softmax}:
\begin{equation}
    \hat{a}=\underset{a}{\arg\max} \int_{\theta}p(a|s,\theta) \, b(\theta) \, d\theta \, .
\end{equation}
Based on this prediction, it determines the most likely next state, i.e., where the human will go next:
\begin{equation}
    \hat{s}=\underset{s'}{\arg\max} \, p(s'|s,\hat{a}) \, .
\end{equation}
In a deterministic environment, this is $\hat{s}=f(s,\hat{a})$, with $f$ being the dynamics.
The robot then communicates this prediction to the human teacher, for instance by gazing in the direction of $\hat{s}$.

\noindent\textbf{Feedback speed.}
If the robot can control the \emph{speed} of feedback (e.g., the speed at which it moves its end effector or gazes), it can use this to convey how \emph{confident} it is in its prediction; slower speeds indicate lower confidence~\cite{Hough_2017,Zhou_2017}. Given a maximum speed of $v_{max}$, we thus set feedback speed as
\begin{equation}
    v = v_{max} \; p(\hat{s}|s,\hat{a})\int_{\theta}p(\hat{a}|s,\theta) \, b(\theta) \, d\theta \, .
\end{equation}

\noindent\textbf{Grounding in our experimental domain. }
To make this more concrete, consider the domain of decluttering: objects need to be sorted appropriately into bins, and only the human knows the correct sorting mechanism. States are locations of objects and bins,
and actions place objects into bins. Every action $a$ corresponds to a particular bin $B_a$, and we assume the environment is deterministic; taking action $a$ in state $s$ means putting the object from location $s$ into $B_a$. $\phi(s,a)$ is a feature vector that consists of descriptors of object-bin match, e.g., distance, color match, shape match, etc.

Each human action teaches the robot about the relationship between the features and the reward, i.e., about the correct weights $\theta^*$. Once the human has selected an object, the robot predicts, according to its current belief, which action $\hat{a}$ the human will take next, and indicates $B_{\hat{a}}$, for instance via gaze. Each $\theta$ assigns different probabilities to different bins, and the robot combines all this information to compute a confidence in its estimate of bin $B_{\hat{a}}$, which it uses to adjust its feedback speed. 

\section{Analysis with Ideal Users: Why Do We Expect Feedback to Help?}
\label{sec:analysis_theory}
\subsection{Model of human teachers that incorporates feedback}
To provide a theoretical justification for feedback, we construct potential models of human teaching that incorporate feedback, and analyze how feedback improves teachers' decisions. These models are based on algorithmic teaching~\cite{Zhu_2015,Goldman_1995,Balbach_2009}, and thus have two components: tracking the learner's state (e.g., what does the learner know), and selecting informative demonstrations based on this.

\noindent\textbf{Tracking learner state.}
We assume our teacher knows how the learner performs an update after every example via \eref{eqn:bayes_irl}. However, the teacher is still missing information about the learner, in particular the learner's \emph{prior} belief $b_0$, and the feature space $\Theta$ that the learner assumes---which may differ from the teacher's. A sophisticated teacher is aware of this \emph{uncertainty}. At every step, she has a belief over what the robot's belief might be, that takes into account the robot's learning updates, its feedback, and the uncertainty over which prior and feature space the robot is using. 

After the first time step, this teacher computes the probability of the robot's new belief given the robot's feedback target $x_0$ and speed $v_0$ for the object she picked up, $s_0$, and the action she took, $a_0$:
\begin{equation}
    p(b_1|s_0,a_0,x_0,v_0)
    =\int_{b_0,\Theta} \underbrace{p(\Theta)p(b_0|\Theta)}_{\text{teacher priors}}\underbrace{p(x_0,v_0|b_0,s_0)}_{\text{feedback generation \eref{eqn:gaze_g_prob}\eref{eqn:gaze_v_prob}}} 
    \underbrace{p(b_1|b_0,s_0,a_0,\Theta)}_{\text{robot learning \eref{eqn:bayes_irl}}} d\Theta \, d b_0 \, . 
\end{equation}
Since we assume the teacher knows how the learner performs belief updates, for the robot learning distribution, $p(b_1|b_0,s_0,a_0,\Theta)$, the teacher places all probability mass on the correct new belief according to \eref{eqn:bayes_irl}. For the feedback generation distribution, $p(x_0,v_0|b_0,s_0)$, we use how the robot actually generates feedback. From $x_0$, the teacher recovers $\hat{a}_0=f^{-1}(s_0,x_0)$, the action that the robot predicted, based on inverse dynamics $f^{-1}$.
From there, the teacher computes the probability of that action under the belief $b_0$:
\begin{equation}
    p(x_0|b_0,s_0)=p(\hat{a}_0|b_0,s_0)=\int_{\theta}p(\hat{a}_0|s_0,\theta) \, b_0(\theta) \, d\theta \, .
    \label{eqn:gaze_g_prob}
\end{equation}
For $v_0$, the teacher assigns probability based on a Gaussian around her predicted speed for $\hat{a}_0$:
\begin{equation}
      p(v_0|b_0,s_0,x_0) =p(v_0|b_0,s_0,\hat{a}_0) =\mathcal{N}(v_0 | v_{max} \; p(\hat{a}_0|b_0,s_0), v_{var}) \, .
    \label{eqn:gaze_v_prob}
\end{equation}
The teacher keeps incorporating such updates based on robot feedback for each new demonstration. In practice, we approximate this update by initializing the teacher's priors $p(\Theta)$ and $p(b_0|\Theta)$ with sampled learner prior beliefs $b_0$ and possible feature spaces $\Theta$, and iteratively updating this set of samples with the evidence, analogous to running a particle filter~\cite{Gordon_1993} (see \sref{sec:feedback_impact} for details).

We also consider a less sophisticated teacher model, who does not account for the uncertainty over the robot's prior and feature space. Rather than maintaining a belief over beliefs, this \emph{iterative} teacher starts with a uniform belief over what the robot's reward estimate might be, and updates it at every step. First, the teacher updates based on the feedback:
\begin{equation}
    b_1'(\theta|s_0,x_0,v_0)\propto p(x_0,v_0|\theta,s_0) \, b_0(\theta) \, ,
\end{equation}
with the feedback probabilities computed as in \eref{eqn:gaze_g_prob} and \eref{eqn:gaze_v_prob}, but for a single $\theta$. 
Next, the teacher shows $a_0$ and accounts for the fact that the robot will learn:
\begin{equation}
    b_1(\theta|s_0,a_0,x_0,v_0)\propto p(a_0|\theta,s_0) \, b_1'(\theta) \, .
\end{equation}
Motivated by the ``win stay, lose shift'' strategy in cognitive psychology~\cite{Nowak_1993}, iterative teachers only update if the feedback disagrees with their current belief; see Appendix \ref{sec:iterative_teacher} for details.

\noindent\textbf{Selecting examples based on current learner state.}
At every step, our teacher uses the tracked learner state to select the most informative example to give next. This is the example that leads to the largest increase in learner performance. The iterative teacher teaches:
\begin{equation}
(s_t^*,a_t^*) = \underset{s_t,a_t}{\arg\max} \, g(b_{t+1}(\theta|s_t,a_t)),
\end{equation}
where $g(\cdot)$ computes the learner's expected performance on the task (if its belief were $b_{t+1}$) and $b_{t+1}(\theta|s_t,a_t) \propto p(a_t|\theta,s_t) \, b_t(\theta)$, according to the learner's learning update via \eref{eqn:bayes_irl}.

The uncertainty-aware teacher does the same, but in expectation over the robot's current belief and the feature space the robot uses:
\begin{equation}
    (s_t^*,a_t^*)= \underset{s_t,a_t}{\arg\max} \, \mathbb{E}_{b_{t},\Theta} \, [ g(b_{t+1}(\theta|s_t,a_t)) ]
\end{equation}

\subsection{Impact of feedback}
\label{sec:feedback_impact}
We now investigate how much feedback helps for our models of human teaching. If the teacher has a perfect model of the learner, then feedback is not necessary~\cite{Balbach_2009}. So, we focus on the kinds of mismatches that might occur---the learner might start off with a bad prior, or not know the correct feature space. For these simulated experiments, we randomly sampled $N$ bins and $O$ objects; each bin and object is represented by a $d$-dimensional feature vector. We set $N = 3$, $O = 50$, and $d = 3$.

\noindent\textbf{Manipulated variables.}
We manipulate the teacher type: \emph{iterative}, \emph{uncertainty-aware}, and \emph{random}. The \emph{random} teacher is a baseline, that chooses random teaching examples rather than maximally informative ones. The learners vary along two factors: whether they provide feedback or not, and whether the teacher-learner mismatch is the prior (\emph{prior mismatch}), the learner missing a feature (\emph{feature mismatch}), or the learner learning a separate $\theta$ for each bin (\emph{reward generalization mismatch}). For details, please refer to Appendix \ref{sec:supp_theoryimpl}. Note that in the latter two mismatch conditions, the learner cannot learn the task, because it is reasoning over the incorrect space of $\theta$s.

\noindent\textbf{Learner performance.}
We measure learning performance with expected soft classification error:
\begin{equation}
    g(b_t(\theta)) = \mathbb{E}_{\theta \sim b_t(\theta)} \left[ \frac{1}{|S|} \sum_{s \in S} p(a^*(s) | s, \theta) \right] \, ,
    \label{eqn:learner_perf}
\end{equation}
where $a^*(s) = \text{argmax}_a \mathcal{R}_{\theta^*}(s, a)$ denotes the correct bin to place the object in.

\begin{figure*}
\centering
\begin{minipage}{.26\linewidth}
  \centering
  \includegraphics[width=1.0\linewidth]{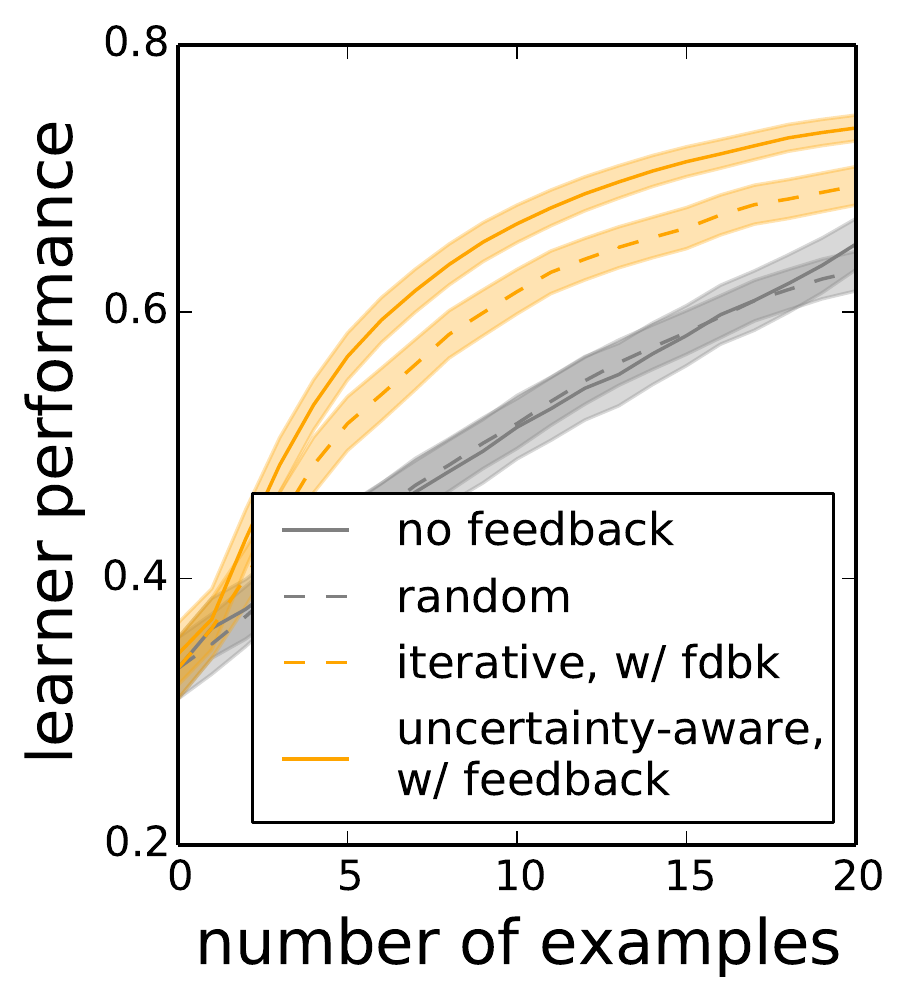}
  \captionof{figure}{For mismatched priors, algorithmic teaching with feedback achieves higher learner performance \eqref{eqn:learner_perf}. Standard error bars are computed from $100$ trials.}
  \label{fig:sim_teaching_effectiveness}
\end{minipage}\enskip%
\begin{minipage}{.72\linewidth}
  \centering
  \includegraphics[width=1.0\linewidth]{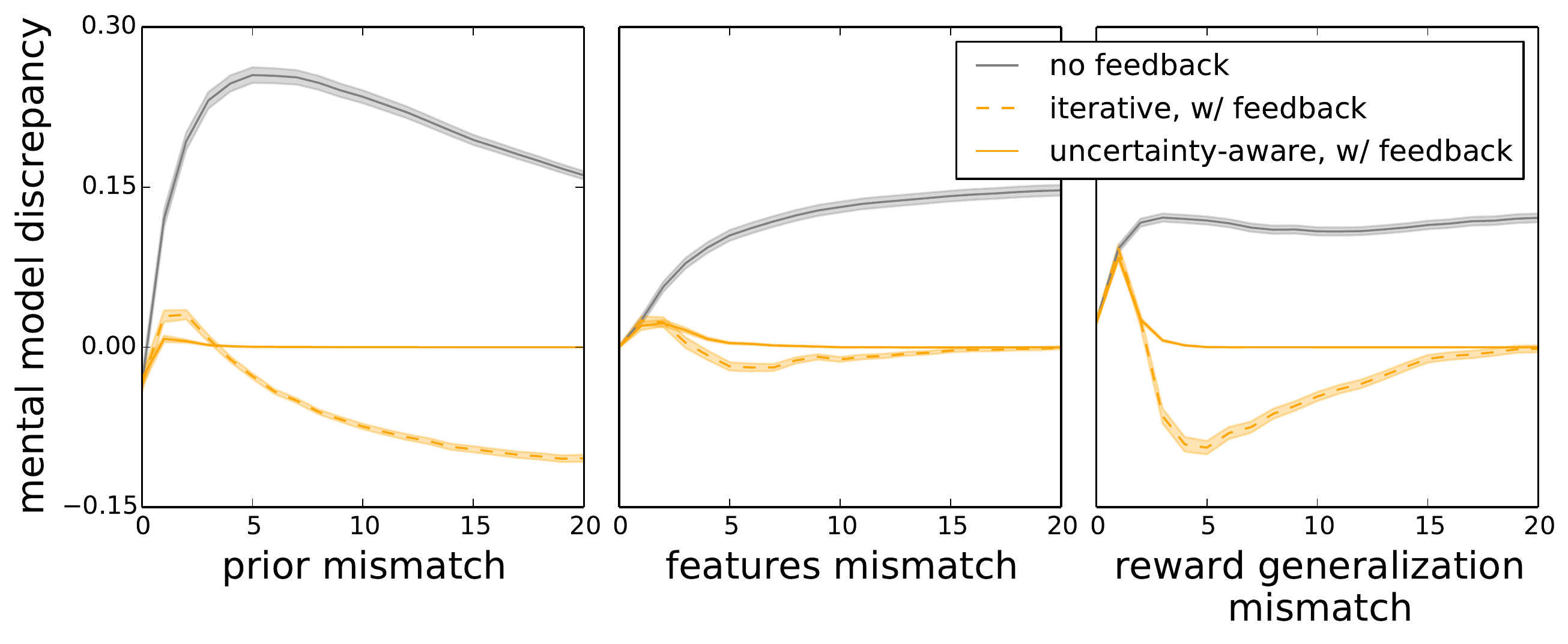}
  \captionof{figure}{With feedback, iterative and uncertainty-aware teachers (orange) can track the learner's state accurately for most mismatch conditions. In contrast, teachers significantly overestimate learner performance when there is no feedback. Standard error bars are computed from 100 trials with ten random teaching sequences each. The y-axis is the teacher's estimate of learner performance minus true performance \eqref{eqn:learner_perf}; the x-axis is the number of teaching examples.}
  \label{fig:sim_teaching_mentalmodel}
\end{minipage}
\vspace{-1.5em}
\end{figure*}

\noindent\textbf{Analysis.}
As one might expect, we found that the effect of feedback depends on the teacher-learner mismatch and teacher type. We found two main positive effects: feedback allows the teacher to 1) select more effective teaching examples, and 2) track the capabilities of the learner more accurately.

Feedback leads to more effective teaching for only the \emph{prior mismatch} condition (\figref{fig:sim_teaching_effectiveness}).
Across all mismatch conditions, uncertainty-aware teachers very quickly narrow down which learner model is correct, and can nearly-perfectly estimate the learner's performance (\figref{fig:sim_teaching_mentalmodel}).\footnote{This makes the potentially strong assumption that the uncertainty-aware teacher's possible learner models contain the true one. This is reasonable though, in the case of \emph{feature mismatch} for interpretable reward features, because then the teacher could just be reasoning over the power set of features.} Although the learner cannot learn the task in the \emph{features mismatch} and \emph{reward generalization mismatch} conditions, there is still a benefit to the teacher estimating the learner's performance correctly: they have a better idea of when to stop teaching (i.e., when their estimated learner performance stops increasing), and they have a more accurate estimate of the learner's performance at the end of training. Uncertainty-aware teachers, because they maintain a belief over what the learner's feature space might be, can also narrow down which feature(s) the learner is missing, which gives them a better idea of the learner's capabilities for future tasks.

\noindent\textbf{Summary.}
Our analysis on algorithmic-teaching-based human models suggests that feedback helps teachers. When the learner can learn the task, feedback makes it easier for teachers to select effective teaching examples. Feedback also improves the teacher's estimate of the learner's capabilities, so when the learner cannot learn the task, feedback enables users to realize this.

\section{Analysis in Practice: Feedback Helps}
\begin{wrapfigure}[17]{R}{0.35\textwidth}
\vspace{-3.5em}
\centering
\frame{\includegraphics[width=.28\columnwidth]{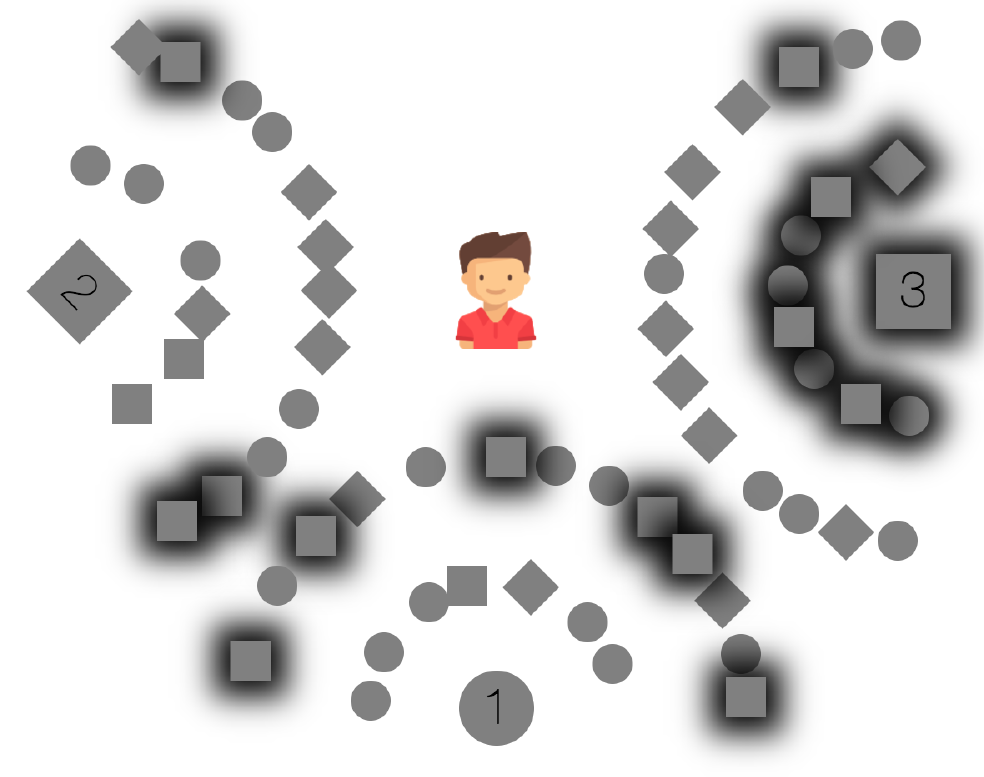}}\\
\vspace{0.4em}
\frame{\includegraphics[width=.28\columnwidth]{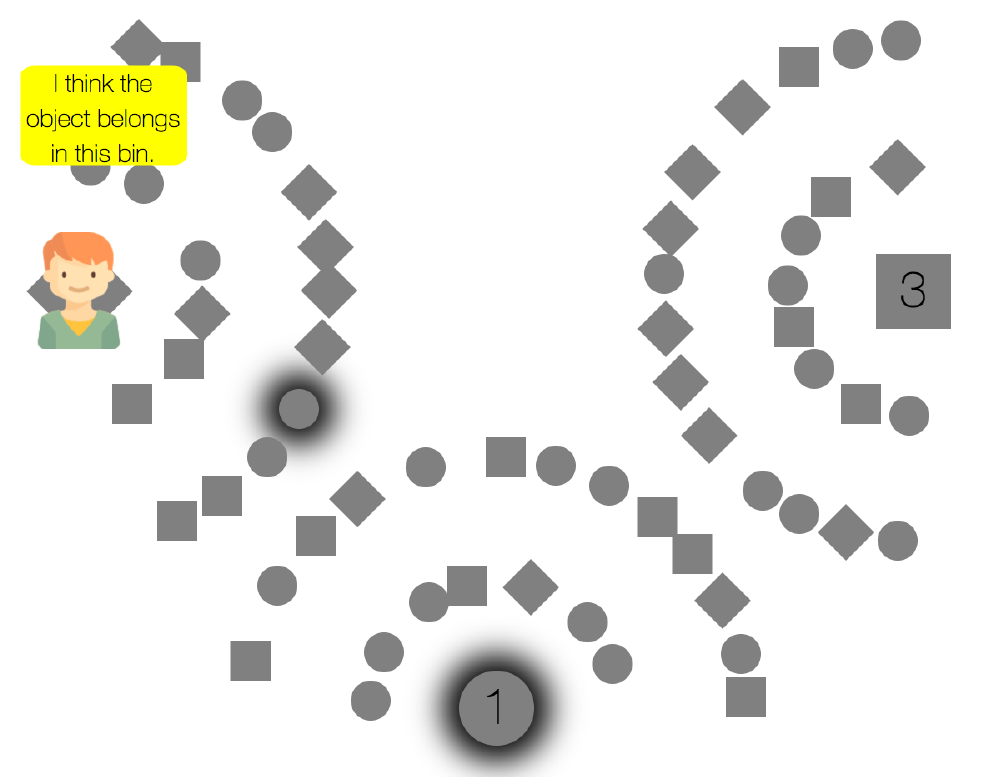}}
\caption{AMT interface. Correct object-bin pairings are shown (top, bin 3), and learners' feedback explicitly signals their prediction (bottom).}
\label{fig:amt_task}
\end{wrapfigure}
We next investigate whether real human teachers also benefit from feedback. In this section, we test the benefits of the feedback itself with virtual learners that explicitly predict the teacher's action with varying confidence. In the next section, we test whether these benefits still exist when the feedback is realized through \emph{gaze} on a real robot.

\subsection{Design}
\label{sec:study_design}
We recruited 87 participants (ages 22-71, 41\% female) on Amazon Mechanical Turk (AMT) to teach a decluttering task to an agent. Participants demonstrated correct object-bin pairings to the agent; we instructed them to give as few teaching examples as possible while still ensuring that the agent learns the correct sorting rules. The decluttering setup (\figref{fig:amt_task}) consisted of three numbered bins surrounded by two rings of objects, with the following sorting rules:
1) an object lying on an inner ring belongs to the closest bin, and 2) an object lying on an outer ring belongs to the bin with the same shape as it.

Users clicked on the object they wanted to demonstrate, and it would be moved into the correct bin. When feedback was activated, the learner avatar would move toward its best guess for the corresponding bin and declare that it thought the object belonged there (\figref{fig:amt_task}, bottom). Then the object would be moved to its bin, and the learner would acknowledge whether they were mistaken.

\noindent\textbf{Manipulated variables.} We manipulated the \emph{feedback}: no feedback, full feedback that indicates the learner's best-guess bin with variable speed corresponding to its confidence level, or partial feedback that indicates the best-guess bin with a fixed speed. We also manipulated \emph{learner prior belief}: a uniform prior the teacher's feature space $\Theta$, a biased prior over $\Theta$ that heavily prefers sorting objects into their closest bin, or a uniform prior over weights in a mismatched feature space different from the teacher's (please refer to Appendix \ref{sec:supp_practiceimpl} for details). We did not conduct the 3 by 3 factorial---instead, we analyze the impact of prior condition separately from the impact of partial feedback, since we did not hypothesize any interactions there. We thus did a 2 by 3 study with full feedback and no feedback, and only tested the partial feedback on the prior belief condition.

\noindent\textbf{Dependent measures.} 
\noindent\textbf{\emph{Objective---}} We measure the \emph{learner performance} \eqref{eqn:learner_perf} throughout teaching. We record the \emph{teaching sequence length} and measure proxy metrics for teaching strategy, e.g., the \emph{proportion of objects demonstrated from each ring}. 
\textbf{\emph{Subjective---}} We also ask open-ended and Likert scale questions about the confidence of the teacher in the learner's understanding, their ability to track the learner's progress, the helpfulness of feedback, and the effects of feedback on teaching. 
\textbf{\emph{Combined---}} In addition, we compute a \emph{mental model discrepancy} metric~\cite{Cakmak_2010b}: the human's Likert estimate of the robot's understanding of a sorting rule (scaled to the 0-1 range), minus the learner performance for objects classified by that rule.

\noindent\textbf{Hypotheses.} We hypothesize that for all learner prior conditions, \textbf{H1}: Feedback will allow the teacher to better track the learner's progress and understanding, \textbf{H2}: The teacher will adapt their strategy according to their improved estimate of learner capabilities, and \textbf{H3}: This adjustment, enabled by feedback, will ultimately result in increased learning performance. 

\noindent\textbf{Subject allocation.}
Participants were randomly allocated between-subjects across the learner prior conditions. Full versus no feedback was within-subjects, because people may vary significantly in terms of their teaching strategies and capabilities, and we wanted to be able to make direct comparisons. We fixed the ordering of conditions: no feedback, followed by feedback. This is because the other ordering would have been too biasing---seeing the feedback first enables participants to figure out how the robot learns, and they can use that information in the no feedback condition if that happens after. Whereas with our ordering, experiencing the no feedback condition first does not inform participants' teaching strategy, because they do not get any signal on what the robot has learned. To mitigate the risk of learning effects, we had participants first complete a practice teaching task and pass a quiz on the sorting rules.

\subsection{Results}
\begin{wrapfigure}[18]{R}{0.5\textwidth}
\vspace{-1.5em}
\centering
\includegraphics[width=0.49\textwidth]{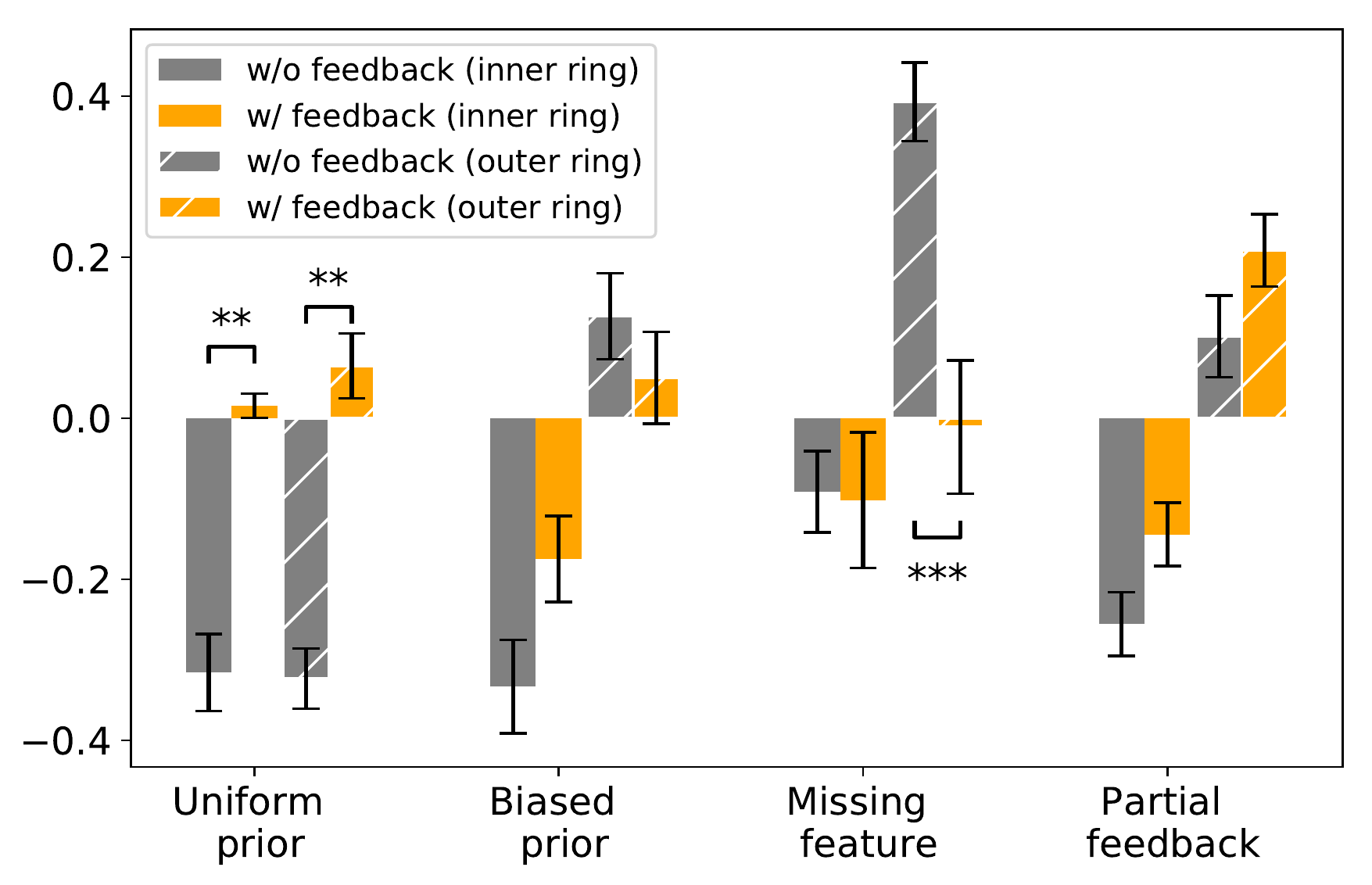}
\caption{Mental model discrepancy (defined in \sref{sec:study_design}) for the inner and outer ring sorting rules. \textbf{\emph{Lower magnitude is better; positive values mean that the teacher overestimates the learner capability, and negative values mean they underestimate.}} (**) indicates $p<0.005$ and (***) $p<0.0005$.}
\label{fig:mental_inner}
\end{wrapfigure}

\noindent\textbf{H1: Feedback improves tracking of robot understanding.} In learner prior conditions with high mental model discrepancy at the end of teaching without feedback, adding feedback reduces this discrepancy (\figref{fig:mental_inner}).
A 2 by 3 factorial repeated-measures ANOVA with feedback and prior belief as factors showed a significant interaction effect on both outer ($F(2,84)=19.78, p<.0001$) and inner ($F(2,84)=4.51, p=0.013)$) ring mental model discrepancy. The post-hoc Tukey HSD found that feedback significantly decreases discrepancy in the uniform prior condition for both the outer ($p=0.001$) and inner ($p=0.002$) ring rules, and in the missing feature condition for outer ($p<0.0001$). Subjective Likert responses also support this hypothesis; see Appendix \ref{sec:supp_likert} for details.

\noindent\textbf{H2: Feedback changes teaching strategy.}
In the biased prior and the missing feature conditions, feedback leads teachers to demonstrate more outer-ring objects (\figref{fig:prop_objects}); an ANOVA found a significant main effect for feedback on this measure ($F(1,85)=31.49, p<.0001$). This is reasonable, since in both conditions, the learner has trouble learning the outer ring rule. An ANOVA on the teaching sequence length found an interaction effect between the feedback and learner mismatch ($F(2, 63)=6.613, p=.0025$). A post-hoc Tukey HSD found that the number of teaching examples is significantly higher for the missing feature case when people receive feedback---they persist in teaching the learner the outer ring rule, until they finally realize it is impossible.

\begin{figure*}[!b]
\vspace{-2em}
\centering
\includegraphics[width=1.0\textwidth]{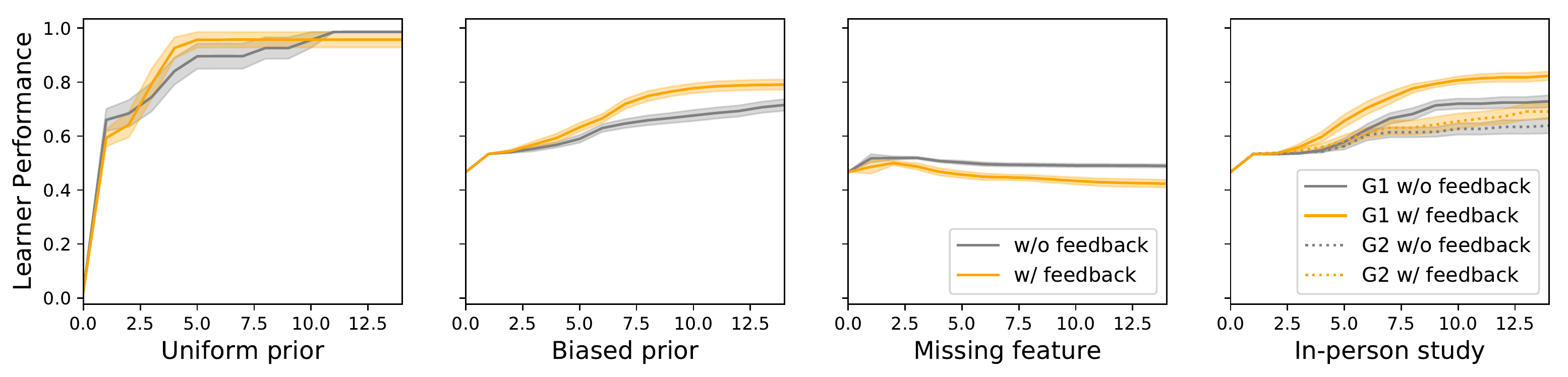}
\caption{Learner performance \eqref{eqn:learner_perf} versus teaching iteration in the online study (left three) and in the in-person study with a biased prior (right-most). The x-axis is the number of teaching examples.}
\label{fig:learner_accuracy}
\end{figure*}

\noindent\textbf{H3: Feedback improves learning performance.}
Feedback during teaching leads to improved learner performance in the biased prior case (\figref{fig:learner_accuracy}), in line with our results on ideal teacher models (\sref{sec:analysis_theory}).
We ran a 2 by 3 factorial repeated-measures ANOVA on final learner performance with feedback and prior belief as factors, and number of examples as a covariate. We found a significant interaction effect between feedback and prior belief ($F(2,1925)=29.74$, $p<.0001$). A post-hoc Tukey HSD found that feedback significantly improves performance for the biased prior case, significantly decreases it for missing feature, and makes no difference for uniform prior. Since the learner cannot learn the task at all in the missing features condition, it is more important that the teacher recognizes the robot's learning limitations---which feedback does help with.

We found no significant differences between full versus partial feedback in terms how effective teachers were (i.e., final learning performance), the proportion of demonstrated objects from the outer ring, and the number of teaching examples shown. However, learners with partial feedback do tend to overestimate learner performance for the outer ring rule (\figref{fig:mental_inner}), possibly because they assume the learner is consistently confident in its prediction.

\noindent\textbf{Summary.} Overall, our results with human teachers were remarkably similar to what we saw with ideal teacher models: 1) feedback improved learning performance in the biased prior condition; and 2) although feedback did not improve performance in the uniform prior or missing feature conditions, it reduced discrepancy between the teacher's model of the learner and the actual learner. 

\section{Analysis in Practice: Robot Gaze}
\begin{wrapfigure}[9]{R}{0.48\textwidth}
\vspace{-3.5em}
\centering
\includegraphics[width=0.47\textwidth]{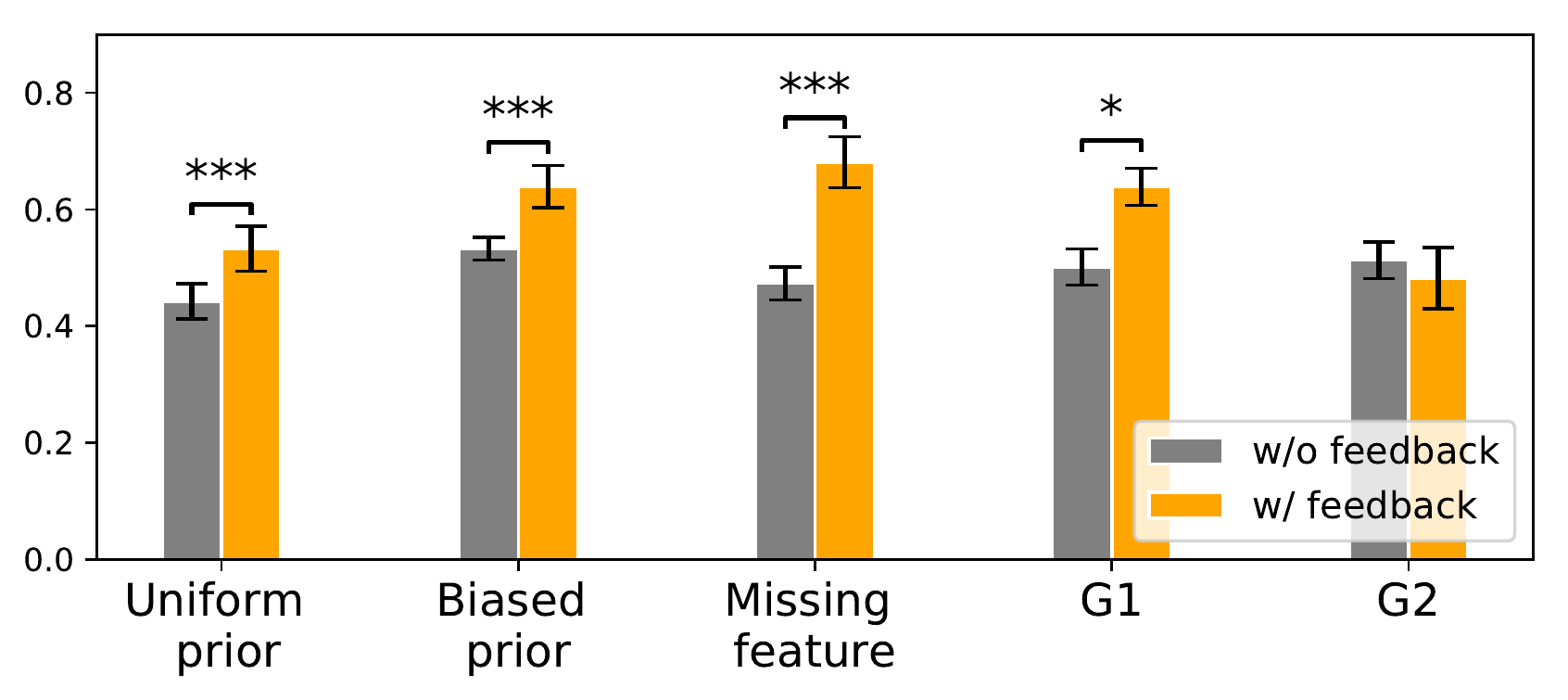}
\caption{Proportion of demonstrated objects from the outer ring on AMT (left) and in-person (right). (*) indicates $p < 0.05$ and (***) $p <0.0005$.} 
\label{fig:prop_objects}
\end{wrapfigure}
We next conducted an in-person study with a PR2 robot. In the real world, gaze cues are more natural and less disruptive than explicit communication. We did not explicitly inform users about the intent of the robot's gaze because we hoped to evaluate whether this choice of nonverbal robot feedback is naturally interpretable---in other words, do humans naturally pick up on gaze as a feedback mechanism for action prediction?

\prg{Design}
We replicated the virtual interface in a real-world decluttering task, using the same layout. Instead of a virtual learner that explicitly moves to its best-guess bin, in this study the PR2 relies on only gaze for feedback, changing its head orientation at varying speeds (Appendix \ref{sec:supp_practice}, \figref{fig:real_task}).

We chose to only test the biased prior condition, because this is where feedback has the largest positive impact on learner performance---for both ideal teacher models and human teachers of virtual learners that provide feedback explicitly. We maintain the same hypotheses from the AMT study.

We recruited 17 users (ages 18-26, 47\% female) from the general student population of our university---a limitation we discuss in \sref{sec:discussion}. Each taught the PR2 the same task twice: first with no feedback, and then with gaze feedback. Participants were told that the robot was reset between the two tasks, so they would not assume the agent retained knowledge from previous demonstrations.

\prg{Results}
Nine participants realized the gaze was related to the robot's belief of the sorting rule. The others misunderstood the gaze as the robot either observing where the demonstrated object was selected from or acknowledging its understanding (since gazing at the bin in front resembles a nod).

We thus report findings separately for those who understood the true intent of the gaze (G1) and those who did not (G2). We ran a repeated-measures ANOVA for each dependent measure, with feedback as a factor. For G1, feedback decreases mental model discrepancy for only the inner ring rule ($F(1, 8) = 16.681, p = 0.0035$), partially supporting \textbf{H1} (\figref{fig:discrepancy_pr2} in Appendix). Subjective Likert responses from G1 also support \textbf{H1}; see Appendix \ref{sec:supp_likert_realworld} for details. Feedback also increases the proportion of examples from the outer ring ($F(1, 8)=9.25, p<.02$) (\figref{fig:prop_objects}), as in the AMT study, and improves learner performance ($F(1,8)=9.53$, $p=.01$) (\figref{fig:learner_accuracy}); this supports hypotheses \textbf{H2} and \textbf{H3}. None of this is true for G2. For both G1 and G2, we found no significant effect of feedback on the teaching sequence length. In addition, virtually no participants were aware of the variations in speed of the robot's head motion, which suggests that more natural and noticeable gaze patterns (e.g., modulating acceleration) should be explored.

\section{Discussion}
\label{sec:discussion}
\noindent\textbf{Summary.}
We find that our proposed form of nonverbal robot feedback, predicting the teacher's next action, helps improve the teacher's effectiveness and mental model of the learner. Findings in practice echo those from our algorithmic teaching model, although we find that communicating confidence does not help significantly in practice. However, it is encouraging that despite our simple implementation of gaze, half of participants naturally interpret it as a feedback mechanism for action prediction.

\noindent\textbf{Limitations and future work.}
Our work is limited in several ways. The main limitation is that our experiments are on a relatively simple, non-sequential task. In the real world, for more complex tasks with larger action spaces, gaze may have less communicative power; investigating effective combinations of gaze and other feedback channels is a promising future direction. In addition, due to the small and biased sample, the results of our study with gaze on the PR2 should be interpreted as trends. Further work is necessary to explore the viability of gaze as a feedback mechanism, e.g., explicitly explaining to users the purpose of gaze, or making the gaze itself more human-like.

Finally, the form of feedback we study only improves learning when the learner can learn the task, but the teacher's model of the learner is not perfect (e.g., a mismatched prior). However, even when the task is not learnable, this form of feedback is still important in helping teachers recognize the robot's limitations and correctly estimate its (lack of) learning.

%===============================================================================

%===============================================================================

%===============================================================================

%===============================================================================

% The maximum paper length is 8 pages excluding references and acknowledgements, and 10 pages including references and acknowledgements

\clearpage
\acknowledgments{This research was funded in part by the National Science Foundation National Robotics Initiative 2.0 (Grant \#1734633) and Cyber-Physical Systems (Grant \#1545126), the Center for Human-Compatible Artificial Intelligence (CHAI), and the Air Force Office of Scientific Research (Grant FA9550-17-1-0308).}
%===============================================================================

\small{
\bibliography{references}  % .bib
}

\normalsize{
\clearpage
\appendix

\section{Analysis with Ideal Users}
\subsection{Model of iterative teachers}
\label{sec:iterative_teacher}
Motivated by the ``win stay, lose shift'' strategy in cognitive psychology~\cite{Nowak_1993}, we model iterative teachers as only updating their belief if the feedback disagrees with their current belief in the learner's state. In other words, if the learner predicts the bin that the teacher's maximum a posteriori (MAP) estimate of the learner's $\theta$ deems most likely, and the corresponding velocity for this $\theta_\text{MAP}$ is within $0.05 * v_\text{max}$ of the learner's feedback speed, then the teacher does not update based on this feedback. Thus, if the teacher has a perfect model of how the learner learns, then whether the learner provides feedback or not, the teacher would maintain the same belief over the learner's hypothesis, and thus teach in the exact same way. If we did not make integrating feedback conditional, then this teacher would potentially teach \emph{worse} in this situation, since the additional feedback-based updates cause the teacher's estimate to diverge from the true one.

\subsection{Implementation details}
\label{sec:supp_theoryimpl}
\noindent\textbf{Learner models.}
In the \emph{prior mismatch} condition, we bias the learner toward a randomly-selected $\theta'$ by setting the prior as the softmax of the distance between $\theta$s:
\begin{equation}
    p(\theta) \propto e^{\beta' \, \theta^\top \theta'},
\end{equation}
with $\beta' = 50$ for a fairly strong bias. In the \emph{feature mismatch} condition, the learner ignores the last feature dimension. In other words, the set of $\theta$s that the learner considers all have $\theta_d = 0$. In the \emph{reward generalization mismatch} condition, the learner learns a separate $\theta$ for each bin, i.e., reward features are computed in terms of only the object $s$, as in $\phi(s)$. So the space of $\theta$s the learner is considering now has dimensions $B*d$.

For learning updates, we set $\beta = 20$ for a learner that assumes demonstrations are close to optimal. For feedback, we set $v_\text{max} = 1.0$ and enforce a minimum speed of $0.05$. We sampled $1000$ $\theta$s to approximate the learner's belief. 

\noindent\textbf{Teacher models.}
In the \emph{prior mismatch} condition, the uncertainty-aware teacher considers $P+1$ possible learner models, one of which is the default (i.e., uniform prior $b_0$); the other $P$ are biased towards different $\theta$s, including the one that the learner is actually biased towards. In the \emph{feature mismatch} condition, the uncertainty-aware teacher considers $d+1$ possible learner models: one with all features, and $d$ that are each missing a different one of the $d$ features. Finally, in the \emph{reward generalization mismatch} condition, the uncertainty-aware teacher considers two possible learner models, one of which has reward features $\phi(s,a)$ that depend on both the object and bin, and the other of which has reward features $\phi(s)$ that depend only on the object. We assume uncertainty-aware teachers start off believing that the learner likely has a uniform prior $b_0$, considers all $d$ features, and uses reward features $\phi(s,a)$---so we set the teachers' priors (i.e., $p(b_0|\Theta)$ and $p(\Theta)$) to place equal probability on all of the models, except ten times more probability on this most-likely learner model. For teacher updates, we assume $v_\text{var} = 0.0025$.

\section{Analysis in Practice}
\label{sec:supp_practice}
\subsection{Participant consent and compensation}
Both the AMT and in-person user studies, along with their corresponding consent forms, received IRB approval. Participants were paid \$3.75 for the AMT study, which lasted around 15 minutes, and participants were given a \$10 Amazon gift card for the in-person study, which lasted just under 20 minutes.

\subsection{Implementation details}
\label{sec:supp_practiceimpl}
We generated 1024 $\theta$s for the learner to reason over, including those that corresponded to conceptually intuitive sorting rules (e.g., objects belong to bins with their same shape). In the uniform prior case, the belief was uniform over these $\theta$s, and in the biased prior case, the learner heavily preferred sorting all objects into their closest bin. In the mismatched features condition, we removed the shape dimension feature in the learner's belief space.

\subsection{Analysis of Likert responses from AMT study}
\label{sec:supp_likert}
\begin{figure*}[t]
    \centering
    \includegraphics[width=0.9\textwidth]{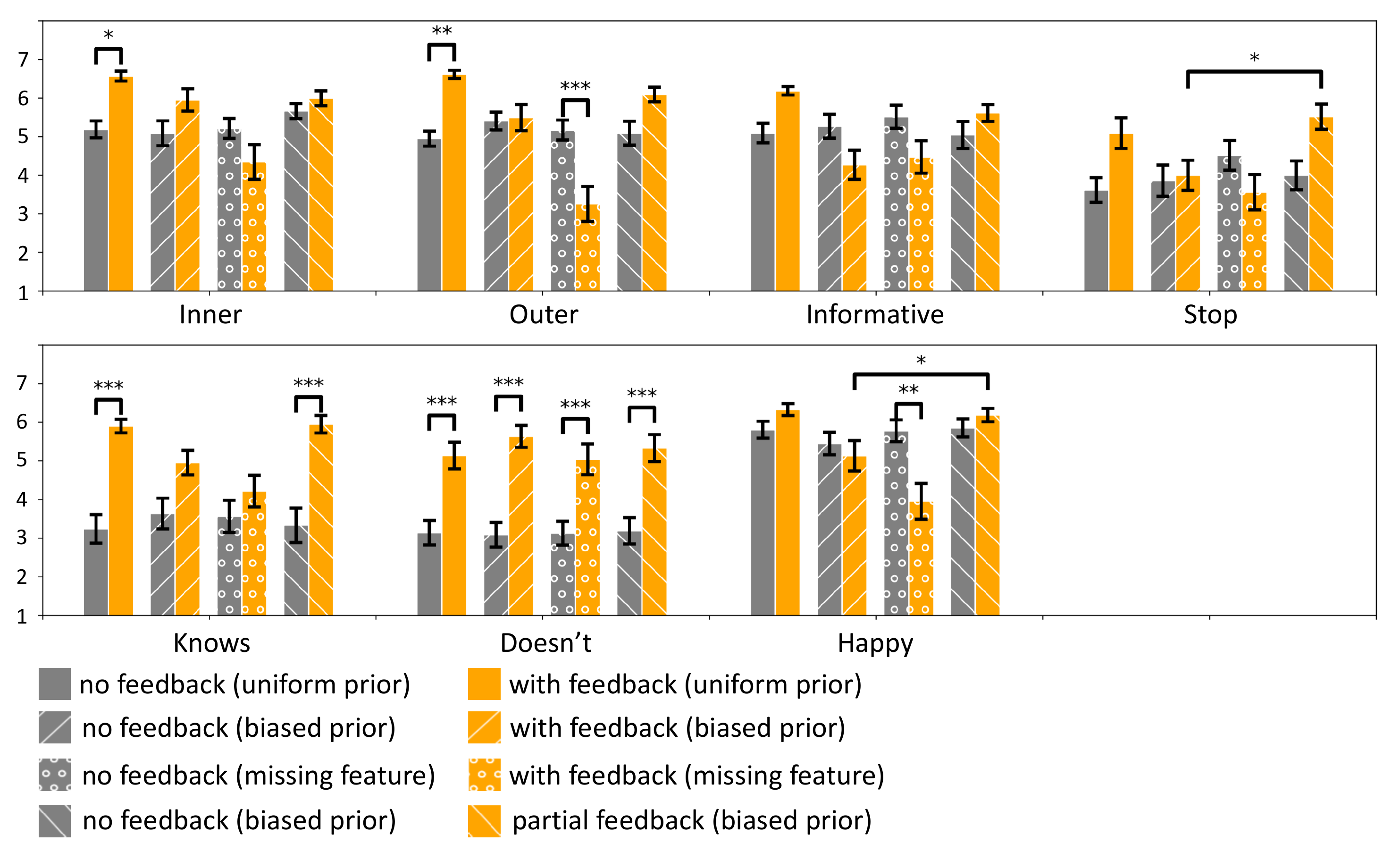}
    \caption{Likert scale responses from the AMT study, in which users taught learners with and without feedback. (*) indicates $p<0.05$, (**) indicates $p<0.005$ and (***) indicates $p<0.0005$. Labels correspond to the Likert questions in Table \ref{tab:turk_likert_anova}.}
    \label{fig:turk_likert}
\end{figure*}

For the AMT study, we ran a two-way repeated measures ANOVA on each of the subjective measures (i.e., Likert questions) with the feedback and prior belief as factors. The full details are reported in Table \ref{tab:turk_likert_anova}. For clarity, in the following analysis we will refer to the questions by their corresponding labels; full questions can be found in the first column of Table \ref{tab:turk_likert_anova}.

\noindent\textbf{Tracking robot understanding.}
In the uniform prior condition, when the learner is able to learn both the inner and outer ring sorting rules, participants' subjective Likert responses indicate that when they were given feedback, they were significantly more confident that the learner learned correctly (\figref{fig:turk_likert}, \emph{inner} and \emph{outer}). A post-hoc Tukey HSD found a significant difference between feedback and no feedback for both \emph{inner} ($p=0.004$) and \emph{outer} ($p=0.004$).

Similarly, in the missing feature condition, when the learner is not able to learn the outer ring rule, participants who received feedback were more aware of this; a post-hoc Tukey HSD found a significant difference between feedback and no feedback for \emph{outer} ($p=0.0003$).

Across all prior belief conditions, participants found it overwhelmingly easier to recognize what the learner \emph{does not} understand; the ANOVA found a significant main effect for feedback for \emph{doesn't}. Participants also generally found it easier to recognize what a learner \emph{knows}; a post-hoc Tukey HSD found a significant difference between feedback and no feedback for the uniform prior ($p<0.0001$) and partial feedback ($p<0.0001$) conditions.

\noindent\textbf{User satisfaction.}
For \emph{happy}, a post-hoc Tukey HSD found a significant difference between feedback and no feedback for the missing feature condition ($p=0.0011$). This makes sense, because when feedback was given in the missing feature condition, participants were clearly aware that the robot was unable to learn the task, despite their efforts to teach it.

\noindent\textbf{Full versus partial feedback.}
An ANOVA on each of the subjective measures with the feedback level as a factor (full feedback versus partial feedback) found significant differences between the partial and full feedback condition for only the \emph{stop} ($F(1, 47)= 8.43, p=0.0059$) and \emph{happy} ($F(1, 47) = 5.57, p=0.02$) questions. The average for the partial feedback condition was higher in both cases. We hypothesize that this is because participants found the partial feedback (with constant speed) to be easier to reason about. This suggests that there is a need for investigating other mechanisms for expressing feedback confidence. %The average for the partial feedback condition was higher in both cases ($5.5$ vs $4$ for \emph{stop} and $6.1$ vs $5.1$ for \emph{happy}).

{\begin{table}[t]
    \small
    \centering
    \begin{tabular}{lccc}
        \toprule \hline \Tstrut
        \textbf{Statement} & \textbf{Effect} & \textbf{F-score} & \textbf{p-value}  \\
        \midrule
        \begin{tabular}[c]{@{}l@{}}\textbf{Inner:} ``[Learner] correctly learned the rule that if an\\ object lies in an inner ring, it belongs to the closest bin''\end{tabular} & interaction & $F(3, 83)=7.99$ & $p<0.0001$\\
        \midrule
        \begin{tabular}[c]{@{}l@{}}\textbf{Outer:} ``[Learner] correctly learned the rule that if an\\ object lies in an outer ring, it belongs to the bin with the\\ same shape''\end{tabular} & interaction & $F(3, 83)=14.12$ & $p<0.0001$\\
        \midrule
        \begin{tabular}[c]{@{}l@{}}\textbf{Informative:} ``I found it easy to choose informative\\ teaching examples for [Learner]''\end{tabular} & none & --- & ---\\
        \midrule
        \begin{tabular}[c]{@{}l@{}}\textbf{Stop:} ``It was easy to know when to stop teaching\\ $$[Learner]''\end{tabular} & none & --- & ---\\
        \midrule
        \begin{tabular}[c]{@{}l@{}}\textbf{Knows:} ``It was easy to tell what [Learner] knows about\\ the task rules''\end{tabular} & interaction & $ F(3, 83)=3.89$ & $p=0.01$\\
        \midrule
        \begin{tabular}[c]{@{}l@{}}\textbf{Doesn't:} ``It was easy to tell what [Learner] \textit{doesn't}\\ know about the task rules''\end{tabular} & interaction & $F(1, 83)=87.7$ & $p<0.0001$\\
        \midrule
        \begin{tabular}[c]{@{}l@{}}\textbf{Happy:} ``I would be happy to teach [Learner] again''\end{tabular} & interaction & $F(3, 83)=6.14$ & $p=0.0008$\\
        \midrule
        \bottomrule
    \end{tabular}
    \vspace{2mm}
    \caption{ANOVA Results for AMT Study Likert Questions.}
    \label{tab:turk_likert_anova}
\end{table}
}

\begin{figure}
\centering     %%% not \center
\subfigure[]{\label{fig:real_a}\frame{\includegraphics[width=.24\columnwidth]{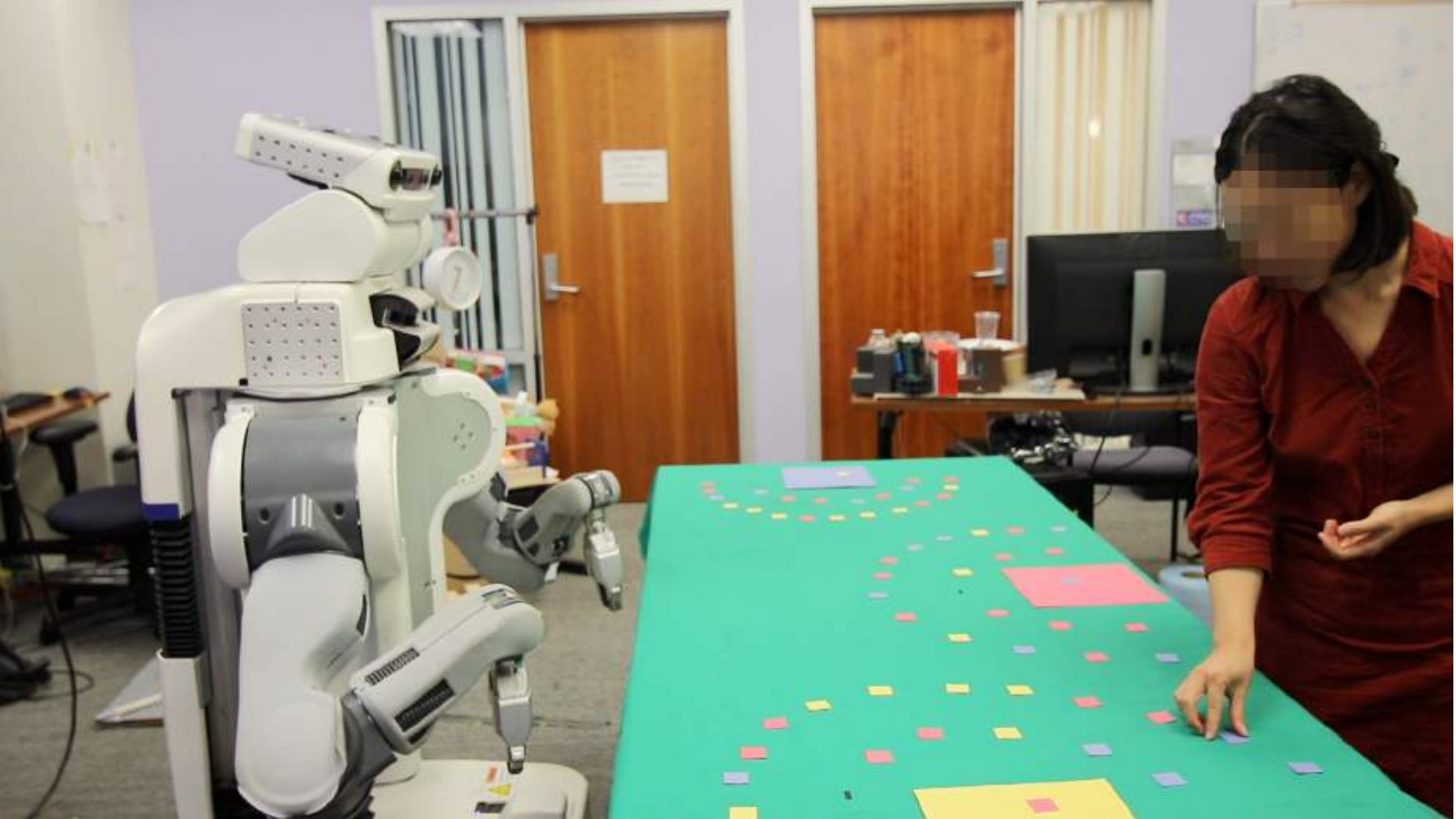}}}
\subfigure[]{\label{fig:real_b}\frame{\includegraphics[width=.24\columnwidth]{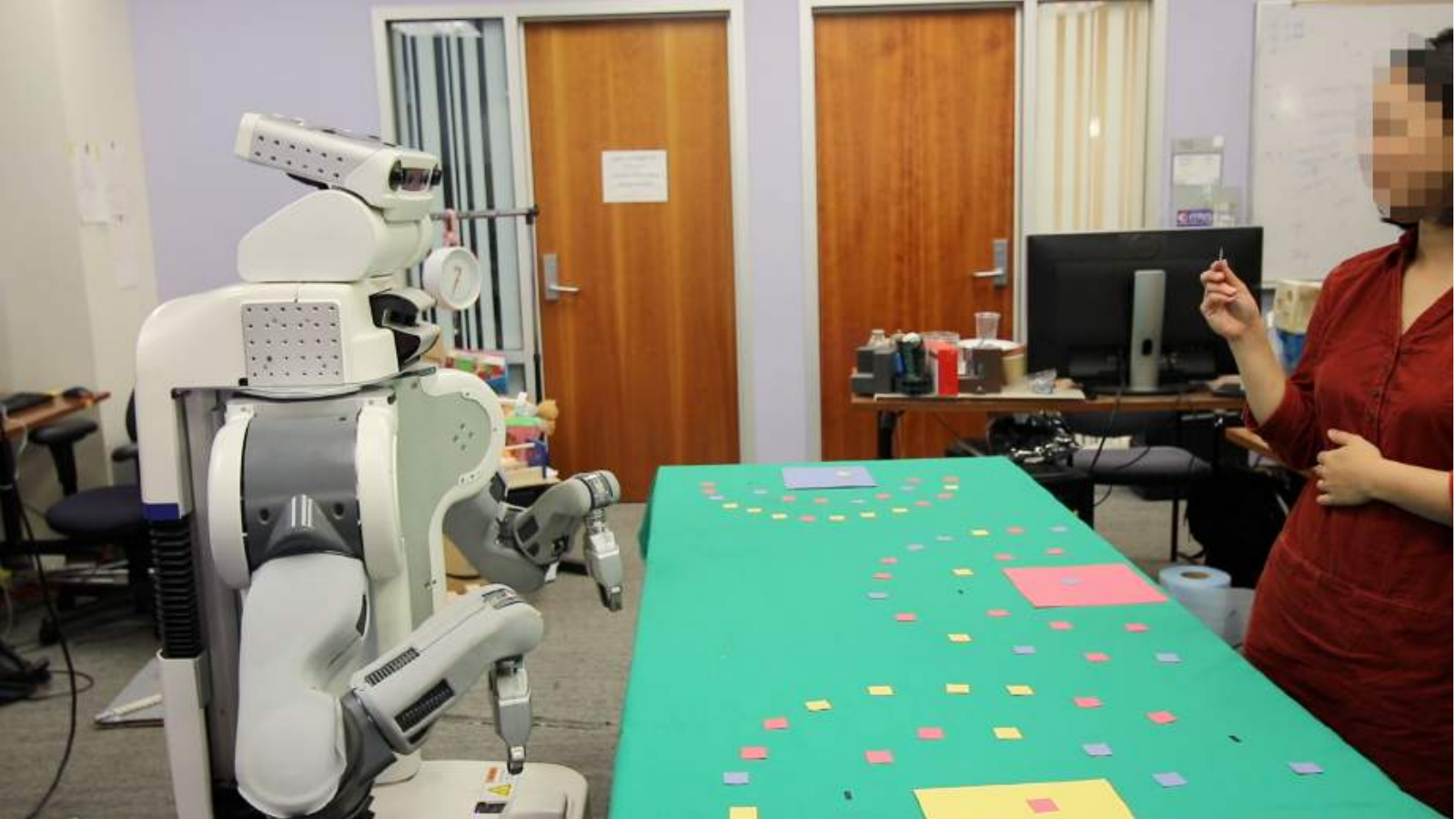}}}
\subfigure[]{\label{fig:real_c}\frame{\includegraphics[width=.24\columnwidth]{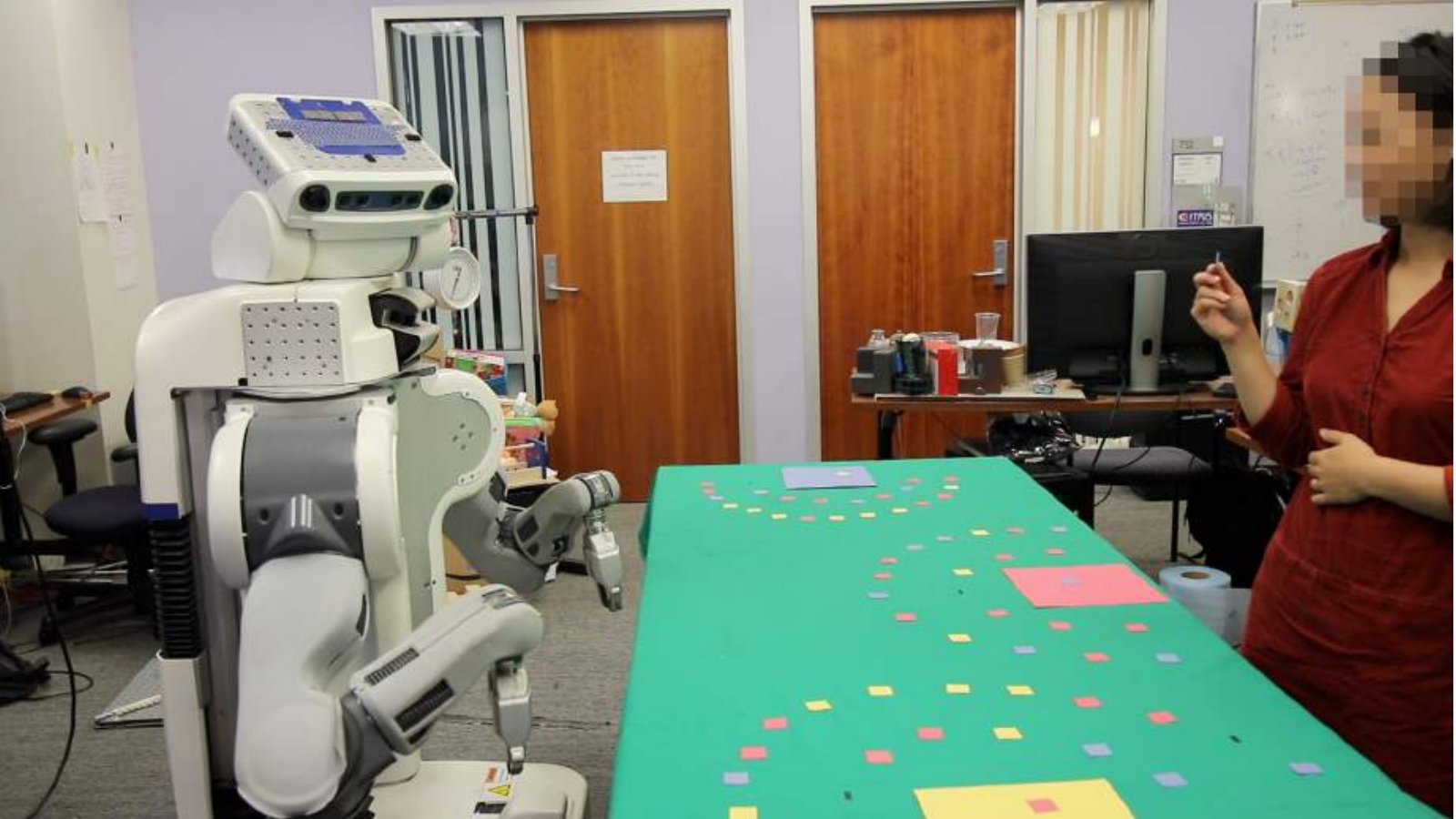}}}
\subfigure[]{\label{fig:real_d}\frame{\includegraphics[width=.24\columnwidth]{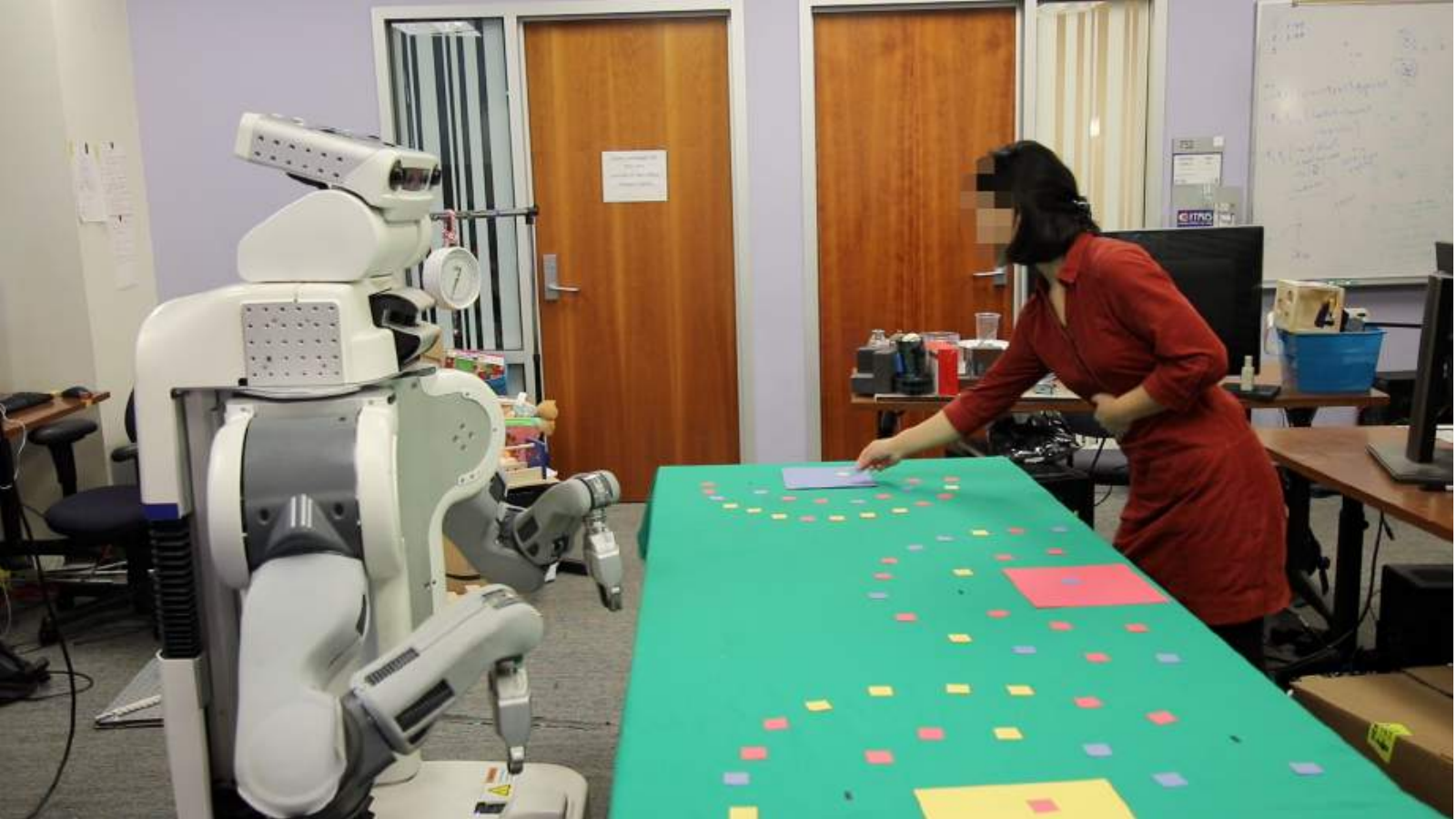}}}
\caption{Real-world experiment flow: the participant a) selects the object to demonstrate, b) picks it up and shows it to the PR2, c) observes the PR2's gaze feedback, and d) places the object in the correct bin.}
\label{fig:real_task}
\end{figure}

\subsection{Analysis of Likert responses from in-person study}
\label{sec:supp_likert_realworld}
\begin{wrapfigure}[17]{R}{0.5\textwidth}
\vspace{-1.5em}
%\begin{figure*}[!h]
%\vspace{-2em}
\centering
\includegraphics[width=0.49\textwidth]{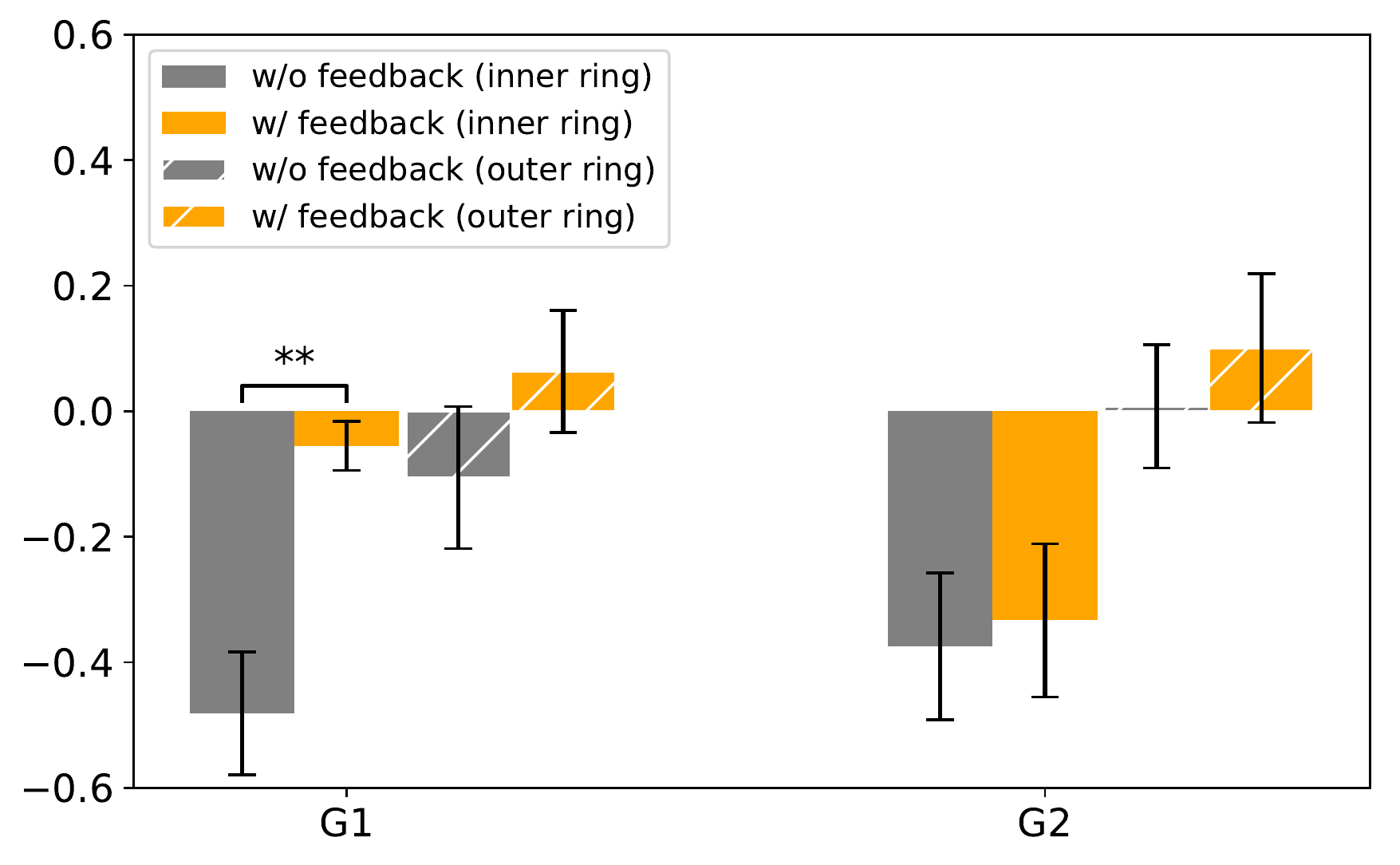}
\caption{Mental model discrepancy (defined in \sref{sec:study_design}) for the inner and outer ring sorting rules. \textbf{\emph{Lower magnitude is better; positive values mean that the teacher overestimates the learner capability, and negative values mean they underestimate.}} (**) indicates $p<0.005$.}
\label{fig:discrepancy_pr2}
%\end{figure*}
\end{wrapfigure}

For the in-person study with the PR2 robot, we ran a repeated-measures ANOVA on each of the subjective measures (i.e., Likert questions) with feedback as a factor. These questions are the same as for the AMT study. The full details are reported in Table \ref{tab:realworld_likert_anova}.

\noindent\textbf{Tracking robot understanding.}
Recall that in this study, the PR2 robot started with a biased prior over weights in $\Theta$, that heavily favors sorting all objects into their closest bin. This biased prior agrees with the inner rule, but contradicts the outer rule. When participants received gaze feedback from the robot, those who understood the true intent of the robot's gaze (G1) better recognized that it correctly learned the \emph{inner} rule ($F(1,8) = 16.66$, $p = 0.0035$), compared to when there was no feedback. Participants in G1 also thought it was easier to give \emph{informative} teaching examples ($F(1,8) = 12.57$, $p = 0.0076$), recognize when to \emph{stop} teaching ($F(1,8) = 14.59$, $p=0.0051$), and recognize what the robot \emph{knows} ($F(1,8) = 139.13$, $p<0.0001$) and \emph{doesn't} know ($F(1,8) = 144.00$, $p < 0.0001$) when the robot provided gaze feedback.

Even participants who misunderstood the intent of the robot's gaze (G2) thought it was easier to recognize when to \emph{stop} teaching ($F(1,7) = 6.66$, $p=0.036$) and recognize what the robot \emph{knows} ($F(1,7) = 5.64$, $p = 0.049$) when the robot provided gaze feedback.

\noindent\textbf{User satisfaction.}
Participants in G1 were more \emph{happy} to teach the robot in the future, when it provided gaze feedback ($F(1,8) = 10.73$, $p = 0.011$).

\begin{figure*}
\vspace{-1em}
\centering
\includegraphics[width=\textwidth]{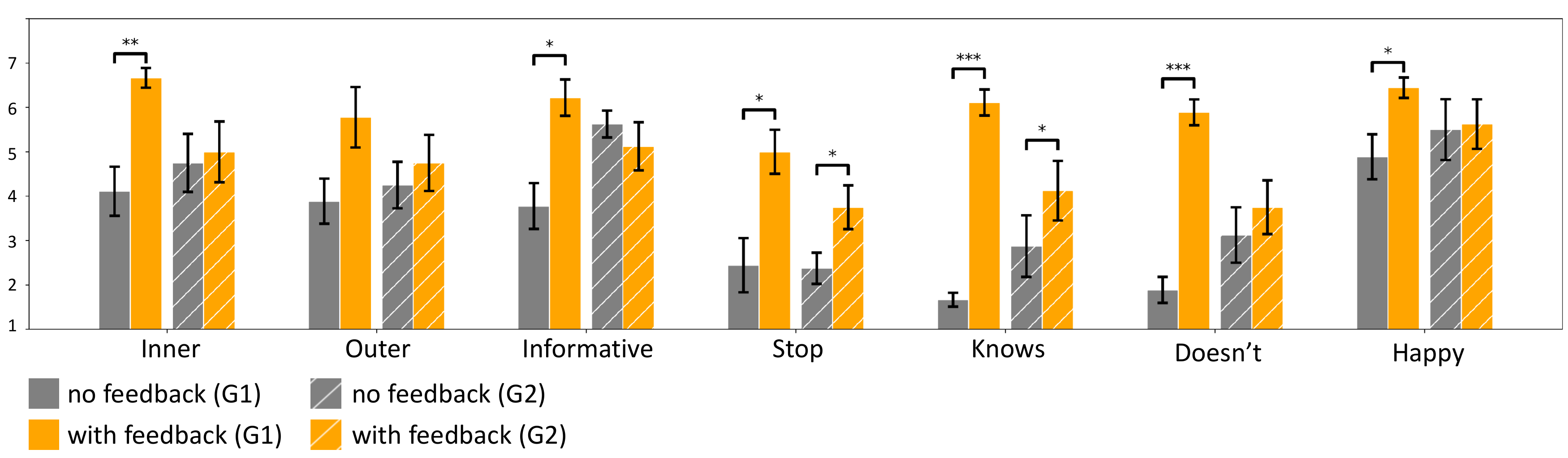}
\caption{Likert scale responses from the in-person study, in which users taught learners with and without feedback. (*) indicates $p<0.05$, (**) indicates $p<0.005$ and (***) indicates $p<0.0005$. Labels correspond to the Likert questions in Table \ref{tab:realworld_likert_anova}.}
\label{fig:realworld_likert}
\end{figure*}

\begin{table}[t]
    \small
    \centering
    \begin{tabular}{lccc}
        \toprule \hline
         \textbf{Statement} & \textbf{Group} & \textbf{F-score} & \textbf{p-value} \Tstrut \\ \midrule

        \rule{0pt}{4ex}\begin{tabular}[c]{@{}l@{}}\textbf{Inner:} ``[Learner] correctly learned the rule that if an\\ object lies in an inner ring, it belongs to the closest bin''\end{tabular} & \begin{tabular}[c]{@{}c@{}} G1\vspace{0.35em}\\G2\end{tabular} & \begin{tabular}[c]{@{}c@{}} $F(1, 8) = 16.66$\vspace{0.35em} \\--- \end{tabular} & \begin{tabular}[c]{@{}c@{}} $p=0.0035$\vspace{0.35em}\\--- \end{tabular} \vspace{0.3em} \\
        \midrule
        \begin{tabular}[c]{@{}l@{}}\textbf{Outer:} ``[Learner] correctly learned the rule that if an\\ object lies in an outer ring, it belongs to the bin with the\\ same shape''\end{tabular} & \begin{tabular}[c]{@{}c@{}} G1\vspace{0.35em}\\G2\end{tabular} & \begin{tabular}[c]{@{}c@{}} ---\vspace{0.35em} \\--- \end{tabular} & \begin{tabular}[c]{@{}c@{}} ---\vspace{0.35em}\\--- \end{tabular} \\
        \midrule
        \rule{0pt}{4ex}\begin{tabular}[c]{@{}l@{}}\textbf{Informative:} ``I found it easy to choose informative\\ teaching examples for [Learner]''\end{tabular} & \begin{tabular}[c]{@{}c@{}} G1\vspace{0.35em}\\G2\end{tabular} & \begin{tabular}[c]{@{}c@{}} $F(1, 8) = 12.57$\vspace{0.35em}\\--- \end{tabular} & \begin{tabular}[c]{@{}c@{}} $p=0.0076$\vspace{0.35em}\\--- \end{tabular} \vspace{0.3em} \\
        \midrule
        \rule{0pt}{4ex}\begin{tabular}[c]{@{}l@{}}\textbf{Stop:} ``It was easy to know when to stop teaching\\ $$[Learner]''\end{tabular} & \begin{tabular}[c]{@{}c@{}} G1\vspace{0.35em}\\G2\end{tabular} & \begin{tabular}[c]{@{}c@{}} $F(1, 8) = 14.59$\vspace{0.35em}\\$F(1, 7) = 6.66$ \end{tabular} & \begin{tabular}[c]{@{}c@{}} $p=0.0051$\vspace{0.35em}\\$p=0.036$ \end{tabular} \vspace{0.3em} \\
        \midrule
        \rule{0pt}{4ex}\begin{tabular}[c]{@{}l@{}}\textbf{Knows:} ``It was easy to tell what [Learner] knows about\\ the task rules''\end{tabular} & \begin{tabular}[c]{@{}c@{}} G1\vspace{0.35em}\\G2\end{tabular} & \begin{tabular}[c]{@{}c@{}} $F(1, 8) = 139.13$\vspace{0.35em}\\$F(1, 7) = 5.64$ \end{tabular} & \begin{tabular}[c]{@{}c@{}} $p < 0.0001$\vspace{0.35em}\\$p=0.049$ \end{tabular} \vspace{0.3em} \\
        \midrule
        \rule{0pt}{4ex}\begin{tabular}[c]{@{}l@{}}\textbf{Doesn't:} ``It was easy to tell what [Learner] \textit{doesn't}\\ know about the task rules''\end{tabular} & \begin{tabular}[c]{@{}c@{}} G1\vspace{0.35em}\\G2\end{tabular} & \begin{tabular}[c]{@{}c@{}} $F(1, 8) = 144.00$\vspace{0.35em}\\--- \end{tabular} & \begin{tabular}[c]{@{}c@{}} $p < 0.0001$\vspace{0.35em}\\--- \end{tabular} \vspace{0.3em} \\
        \midrule
        \rule{0pt}{4ex}\begin{tabular}[c]{@{}l@{}}\textbf{Happy:} ``I would be happy to teach [Learner] again''\end{tabular} & \begin{tabular}[c]{@{}c@{}} G1\vspace{0.35em}\\G2\end{tabular} & \begin{tabular}[c]{@{}c@{}} $F(1, 8) = 10.73$\vspace{0.35em}\\--- \end{tabular} & \begin{tabular}[c]{@{}c@{}} $p=0.011$\vspace{0.35em}\\--- \end{tabular} \vspace{0.3em} \\
        \midrule
        \bottomrule
    \end{tabular}
    \vspace{2mm}
    \caption{ANOVA Results for In-Person Study Likert Questions.}
    \label{tab:realworld_likert_anova}
\end{table}
}

\end{document}